%% file: main_eng.tex
\documentclass[preprint,12pt]{elsarticle}
\usepackage{natbib}
\usepackage{amssymb}
\usepackage{amsmath}
\usepackage{graphicx}
\usepackage{algorithm}
\usepackage{algpseudocode}
\usepackage{array}   
\usepackage{booktabs} 
\usepackage{microtype}
\biboptions{authoryear}

\journal{Information Processing \& Management}

\begin{document}

\begin{frontmatter}

\title{Does AI Homogenize Student Thinking? A Multi-Dimensional Analysis of Structural Convergence in AI-Augmented Essays}

\author[1,2]{Keito Inoshita\corref{cor1}}
\ead{inosita.2865@gmail.com}
\author[3]{Michiaki Omura}
\ead{omura@arbege.com}
\author[4]{Tsukasa Yamanaka}
\ead{yaman@fc.ritsumei.ac.jp}
\author[5]{Go Maeda}
\ead{maeda-g@st.ritsumei.ac.jp}
\author[5]{Kentaro Tsuji}
\ead{k-tsuji@st.ritsumei.ac.jp}

\cortext[cor1]{Corresponding author.}

\affiliation[1]{organization={Ritsumeikan Global Innovation Research Organization (R-GIRO), Ritsumeikan University},
            addressline={1-1-1, Nojihigashi},
            city={Kusatsu},
            postcode={525-8577},
            state={Shiga},
            country={Japan}}

\affiliation[2]{organization={Faculty of Business and Commerce, Kansai University},
            addressline={3-3-35, Yamatecho}, 
            city={Suita},
            postcode={564-8680}, 
            state={Osaka},
            country={Japan}}

\affiliation[3]{organization={Arbege Corporation},
            addressline={3-20, Yotsuyatori}, 
            city={Nagoya},
            postcode={464-0819}, 
            state={Aichi},
            country={Japan}}

\affiliation[4]{organization={College of Life Sciences, Ritsumeikan University},
            addressline={1-1-1, Nojihigashi},
            city={Kusatsu},
            postcode={525-8577},
            state={Shiga},
            country={Japan}}

\affiliation[5]{organization={Office of General Education, Ritsumeikan University},
            addressline={1-1-1, Nojihigashi},
            city={Kusatsu},
            postcode={525-8577},
            state={Shiga},
            country={Japan}}

\begin{abstract}
While AI-assisted writing has been widely reported to improve essay quality, its impact on the structural diversity of student thinking remains unexplored. Analyzing 6,875 essays across five conditions (Human-only, AI-only, and three Human+AI prompt strategies), we provide the first empirical evidence of a Quality-Homogenization Tradeoff, in which substantial quality gains co-occur with significant homogenization. The effect is dimension-specific: cohesion architecture lost 70--78\% of its variance, whereas perspective plurality was diversified. Convergence target analysis further revealed that AI-augmented essays were pulled toward AI structural patterns yet deviated significantly from the Human--AI axis, indicating simultaneous partial replacement and partial emergence. Crucially, prompt specificity reversed homogenization into diversification on argument depth, demonstrating that homogenization is not an intrinsic property of AI but a function of interaction design.
\end{abstract}




\end{frontmatter}

\input{Authors/documents/1_introduction}
\input{Authors/documents/2_related-work}
\input{Authors/documents/3_methodology}
\input{Authors/documents/4_experiment-and-analysis}
\input{Authors/documents/5_discussion}
\input{Authors/documents/6_conclusion}

\section*{Declaration of competing interest}
The authors declare that they have no known competing financial interests or personal relationships that could have appeared to influence the work reported in this paper.

\section*{Acknowledgment}
This work was supported by a research grant from the Ritsumeikan Global Innovation Research Organization (R-GIRO), Ritsumeikan University. The authors also acknowledge the RaaS (Ritsumeikan AI-powered Assessment Solution) system developed and deployed at Ritsumeikan University, which provided practical insights into large-scale AI-assisted assessment (https://raas.ritsumei.ac.jp/).

\section*{Data availability}
The AIDE dataset used in this study is publicly available at Kaggle (https://www.kaggle.com/datasets/lburleigh/tla-lab-ai-detection-for-essays-aide-dataset). The generated essays, extracted structural features, and analysis code will be made available upon publication.

\section*{Declaration of Generative AI Use}
During the preparation of this manuscript, the authors used generative AI tools for language editing and organization. All content was reviewed and verified by the authors, who take full responsibility for the final manuscript.

\bibliographystyle{elsarticle-harv}
\bibliography{refs}

\end{document}

%% file: Authors/documents/1_introduction.tex

\section{Introduction}
\label{sec:introduction}

\subsection{What Has Generative AI Changed About Assessment?}
\label{subsec:what_changed}

The integration of generative AI into higher education has precipitated a crisis of assessment reliability. Across institutions worldwide, the extent to which essays and written assignments can continue to serve as valid measures of student learning has been called into question~\citep{1}. The dominant discourse frames this problem in terms of academic integrity: students can now use AI to produce essays of a quality they could not achieve on their own, thus rendering conventional assessment meaningless~\citep{2}.

While this framing is understandable, it obscures a more fundamental concern. The real crisis is not that students are committing misconduct with AI. Rather, our very conception of competence rests on an implicit assumption that has quietly collapsed: the quality of a written product directly reflects the cognitive ability of the individual who produced it. Although this equation has never been explicitly stated in assessment theory, it has implicitly underpinned every rubric and every grading criterion.

Generative AI has fundamentally altered the conditions under which this assumption holds. Conventional ICT tools such as calculators, search engines, and databases also supported cognition, but what they changed was the input to cognition: what information one could access. The scope of inquiry expanded, yet the act of reasoning with that information remained internal to the individual. The change brought about by generative AI is qualitatively different. AI intervenes not in the input to cognition but in the cognitive process itself, reasoning, argumentation, and the structuring of written discourse. In effect, cognitive boundary permeability has extended from the input layer to the process layer~\citep{3}. This represents a discontinuous shift, not a mere extension of prior ICT use.

It is important to note that this shift does not imply cognitive opening in the autopoietic sense~\citep{4}. The cognitive system remains operationally closed: the human agent continues to interpret, select, and integrate AI outputs. What has changed is not whether the system is open or closed, but the type and depth of information that permeates its boundary. No longer limited to raw materials such as data and facts, processed cognitive products such as inferences, structures, and arguments now flow across the boundary.

The cognitive system that produces a contemporary essay is no longer a single human intellect. In this study, we conceptualize this Human+AI writing system as an Augmented Cognitive Unit ($\mathrm{ACU} = f(\mathrm{Human}, \mathrm{AI}, \mathrm{Interaction})$). The interaction term is not merely additive. How a human designs a prompt and how they integrate AI output can give rise to emergent properties that neither human nor AI would produce in isolation~\citep{5}. At the same time, however, AI patterns may overwrite human structural choices. Whether an ACU achieves genuine cognitive augmentation or merely produces homogenized cognitive patterns is an empirical question.

\subsection{The Unexplored Problem of Homogenization}
\label{subsec:homogenization_problem}

Much of the existing literature on AI and student writing has focused on a single dimension: quality. Do AI-assisted essays receive higher scores? Is grammar more accurate? Are arguments better developed? The general finding is unsurprising: AI augmentation tends to improve quality metrics.

However, this disproportionate focus on quality has entirely overlooked a critically important problem: homogenization, the possibility that AI use causes students' thinking patterns to converge toward the same mold. Quality asks, ``How good is this essay?'' Homogenization asks, ``How similar has this essay become to every other student's essay?'' These are fundamentally different questions, and the answer to one does not predict the answer to the other. Even if quality improves, if every student writes in the same structural pattern, the educational implications are serious. Quality improvement and structural homogenization can co-occur, yet the latter is in principle undetectable under conventional frameworks that evaluate only quality.

The importance of this problem extends beyond a sentimental concern about lost individuality. If we view intellectual activity as an ecosystem, a healthy ecosystem is not one in which everyone achieves maximum quality, but one in which diverse thinking patterns coexist. Just as monoculture in agriculture maximizes short-term yield while introducing long-term fragility, the homogenization of thought may undermine the very preconditions for knowledge creation, critical inquiry, and innovation.

To date, no study has systematically examined whether student writing becomes homogenized across the three conditions that define writing in the age of AI augmentation: Human-only (H), AI-only (A), and Human+AI (H+AI). This three-condition comparison is essential because it can reveal not only whether homogenization occurs, but where homogenized output converges. If H+AI converges near A, it indicates replacement of human thinking; if it falls midway between H and A, it suggests blending; if it occupies a region distinct from both, it points to emergence. Moreover, no study has examined how the mode of interaction with AI, that is, the prompt strategy, moderates the degree of homogenization. A student who pastes an essay and asks the AI to improve it is engaged in a fundamentally different cognitive act than one who instructs the AI to strengthen the argument structure. If the degree of homogenization varies with prompt type, then the problem lies not with AI itself but with how AI is used.

\subsection{The Homogenization Hypothesis and Research Questions}
\label{subsec:hypothesis_rq}

The foregoing discussion leads to the central hypothesis of this study. The Homogenization Hypothesis posits that AI augmentation improves the average quality metrics of student writing while simultaneously homogenizing the structural patterns of thinking across the student population, and that the degree of homogenization varies by structural dimension and interaction mode.

It should be emphasized that this hypothesis does not claim that homogenization is invariably undesirable. Homogenization on certain dimensions, for example, raising the floor of logical cohesion, may be educationally beneficial. The problem is that it is currently unknown on which dimensions homogenization occurs and on which it does not. Without such a map, it is impossible to meaningfully debate whether AI use should be promoted or restricted. This study represents the first attempt to draw that map.

When homogenization is observed, the convergence target carries distinct theoretical implications. If H+AI converges toward A, AI is replacing rather than augmenting student thinking. If it falls between H and A, contributions from both are being diluted. If it occupies a region distinct from both H and A, the human--AI combination is generating novel structures. These three scenarios may manifest differently across dimensions within a single dataset.

This study addresses the following four research questions:

\begin{description}
  \item[RQ1:]  Does AI augmentation improve essay quality while simultaneously homogenizing the structural patterns of student thinking?
  \item[RQ2:] Which structural dimensions of writing are most susceptible to AI-driven homogenization, and which are resistant?
  \item[RQ3:] When homogenization is observed, do the structural profiles of H+AI essays converge toward AI-only output, toward the midpoint between H and A, or toward a structurally novel region?
  \item[RQ4:] Do interaction modes moderate the degree and pattern of homogenization?
\end{description}

The main contributions of this study are threefold:

\begin{enumerate}[i)]
  \item A Quality-Homogenization Tradeoff framework that independently evaluates AI augmentation effects along two axes, quality and structural diversity, and provides the first empirical evidence that quality improvement and homogenization co-occur.
  \item A homogenization map that identifies, for the first time, which structural dimensions are susceptible to and which are resistant to AI-driven homogenization, demonstrating that the effect does not proceed uniformly but varies by dimension.
  \item A prompt moderation analysis that demonstrates that even with the same AI model, prompt design can reverse homogenization into diversification, establishing that homogenization is not an intrinsic property of AI but a function of interaction design.
\end{enumerate}

The remainder of this paper is organized as follows. Section~\ref{sec:related_work} reviews related work. Section~\ref{sec:method} describes the research methodology. Section~\ref{sec:results} reports the results for each of the four research questions. Section~\ref{sec:discussion} discusses theoretical and practical implications along with limitations. Section~\ref{sec:conclusion} presents our conclusions.

%% file: Authors/documents/2_related-work.tex

\section{Related Work}
\label{sec:related_work}

\subsection{AI-Assisted Writing and Its Effects on Quality}
\label{subsec:ai_writing}

Since the introduction of generative AI into education, empirical studies examining the effects of Large Language Models (LLM)-based feedback and co-writing on student writing quality have accumulated rapidly. The overall finding is consistent: AI augmentation tends to improve essay quality metrics.

\citet{6} compared the feedback quality of ChatGPT and trained human raters on 200 secondary school essays and found that human raters outperformed ChatGPT on four of five criteria, although ChatGPT achieved comparable performance on criterion alignment. \citet{7} reported a Randomized Controlled Trial (RCT) with 150 Chinese university students in which the ChatGPT group achieved significantly higher writing scores than both the automated writing evaluation group and the control group. \citet{8} conducted an RCT with 259 university students and confirmed that AI feedback had particularly strong effects on organization ($\beta = 0.311$) and content ($\beta = 0.191$). \citet{9} demonstrated in an RCT with 459 German secondary students that GPT-3.5-turbo-generated feedback significantly improved text revision, learning motivation, and positive affect. \citet{10} reported that, with appropriate guidance, graduate students using generative AI reduced writing time by 56.7\% while improving quality from A$-$ to A.

In addition to these studies, the field of Automated Essay Scoring (AES) has a long research history. Early AES systems relied on handcrafted features such as lexical diversity and syntactic complexity~\citep{11}, but as \citet{12} noted, these systems suffered from a fundamental limitation: excessive dependence on surface-level features. Since 2016, following the introduction of neural approaches by \citet{13}, the application of LLMs such as GPT-4 to AES has progressed rapidly~\citep{14}.

However, all of the studies reviewed above focus on a single dimension, how good an essay is. As \citet{7} suggested through qualitative data, the ChatGPT group exhibited a paradox: writing scores improved while the ideal L2 writing self declined, indicating that quality gains may be accompanied by other, less visible changes. Yet, to the best of our knowledge, no study has systematically measured the impact of AI augmentation on the structural diversity of writing across students.

\subsection{Reduced Output Diversity and Text Homogenization in LLMs}
\label{subsec:homogenization}

A separate body of research has raised concerns about the output diversity of LLMs themselves. \citet{15} reported a controlled experiment in which argumentative essays were written under three conditions: a base LLM, a feedback-tuned LLM (InstructGPT), and humans alone. Co-writing with InstructGPT increased inter-author similarity and significantly reduced both lexical and content diversity. This study constitutes one of the first empirical demonstrations that co-writing with LLMs induces content homogenization and serves as a direct antecedent to the present work.

\citet{16} analyzed 2,200 college admissions essays and showed that human-written essays contributed more novel ideas with each additional essay than GPT-4-generated ones. Despite attempts to enhance AI output diversity through prompt and parameter modifications, the findings suggested that widespread LLM use may reduce creative diversity at the population level. \citet{17} demonstrated in an idea generation experiment using LLM-based creativity support tools that while ChatGPT users generated more ideas at the individual level, group-level homogenization increased. This homogenization was attributed to a group-level phenomenon in which the LLM proposed similar ideas to different users.

Homogenization at the lexical level has also been confirmed across multiple studies. \citet{18} showed that ChatGPT-3.5 used fewer unique words and exhibited lower lexical diversity than humans. \citet{19} proposed a comprehensive evaluation framework encompassing lexical, syntactic, and semantic diversity, and found that instruction tuning improves lexical diversity while constraining syntactic and semantic diversity. \citet{20} found that LLM-generated narratives frequently contained repeated plot elements, quantitatively demonstrating lower story-level originality compared to human writing. \citet{21} demonstrated the risk of model collapse, in which recursive training on synthetic data leads to a consistent decline in output diversity.

These studies make important contributions by showing that LLM output is less diverse than human text. However, the existing literature has several critical gaps. First, most of the studies reviewed above compare only AI-only output with human-only output; how AI-augmented human text behaves in terms of diversity patterns has not been examined. Second, diversity measurement has been limited primarily to the lexical level or the idea level, and diversity in structural dimensions of essays, such as argument depth, cohesion architecture, and structural originality, has not been measured. Third, no study has examined how differences in prompt strategy alter homogenization patterns. The present study addresses these three gaps by conducting, within a single design, a three-condition comparison (H/A/H+AI), variance measurement across six structural dimensions, and comparison across three prompt conditions.

\subsection{Validity of LLM-Based Essay Evaluation}
\label{subsec:llm_evaluation}

Because our methodology uses GPT-5~\citep{22} for structural feature extraction, it is necessary to review prior work on the validity of LLM-based text evaluation. \citet{23} systematically examined the LLM-as-a-Judge paradigm and showed that GPT-4 as a judge achieved over 80\% agreement with human preferences, comparable to inter-rater agreement among human evaluators. At the same time, they identified limitations including position bias, verbosity bias, and self-enhancement bias. \citet{24} examined the reliability of multi-dimensional essay evaluation using GPT-4 and reported high accuracy under criterion- and sample-referenced prompts, with particularly strong performance on idea and organization, though performance on conventions was limited.

In essay scoring applications, \citet{25} confirmed high intra-rater reliability and moderate positive correlation with human scores when GPT-4 was used to score 300 university English placement essays. \citet{26} reported that GPT-4 evaluation of dental student essays achieved excellent inter-rater agreement. \citet{27} found that a fine-tuned GPT-3.5 achieved 78.3\% exact agreement with human raters, although it tended to overrate lower-level essays.

At the same time, biases in LLM-based evaluation warrant careful attention. \citet{28} quantitatively demonstrated that GPT-4 assigns significantly higher scores to texts with lower perplexity, indicating a self-preference bias. \citet{29} systematically classified 12 types of potential bias including position, verbosity, and self-enhancement bias, and showed that even advanced models exhibit notable biases on specific tasks. \citet{30} reported that LLM-based AES tends to rate GPT-generated text more favorably than human-written text, highlighting the bias problem in AES in the era of generative AI.

Taken together, these findings indicate that LLM-based evaluation achieves moderate-to-high agreement with human judgment for holistic quality scoring, while posing bias risks particularly when evaluating AI-generated text. In the present study, we mitigate the influence of such biases by focusing not on absolute quality scores but on between-condition variance changes as the primary analytic target, combined with stability verification through three independent extraction runs (CV $< 0.10$).

%% file: Authors/documents/3_methodology.tex

\section{Method}
\label{sec:method}

\subsection{Research Design}
\label{subsec:design}

This study examines the impact of AI augmentation on the structural diversity of student writing. The core methodological innovation lies in comparing not only quality but also the distributional properties of structural features across five essay conditions: H, A, and three Human+AI augmentation conditions with varying prompt strategies (H+AI Minimal, H+AI Structural, H+AI Delegative). By measuring the structural characteristics of each essay as a six-dimensional vector, the design enables independent assessment of quality improvement and structural homogenization, two phenomena that have been conflated under conventional single-score evaluation.

\begin{figure}[t]
  \centering
  \includegraphics[width=\textwidth]{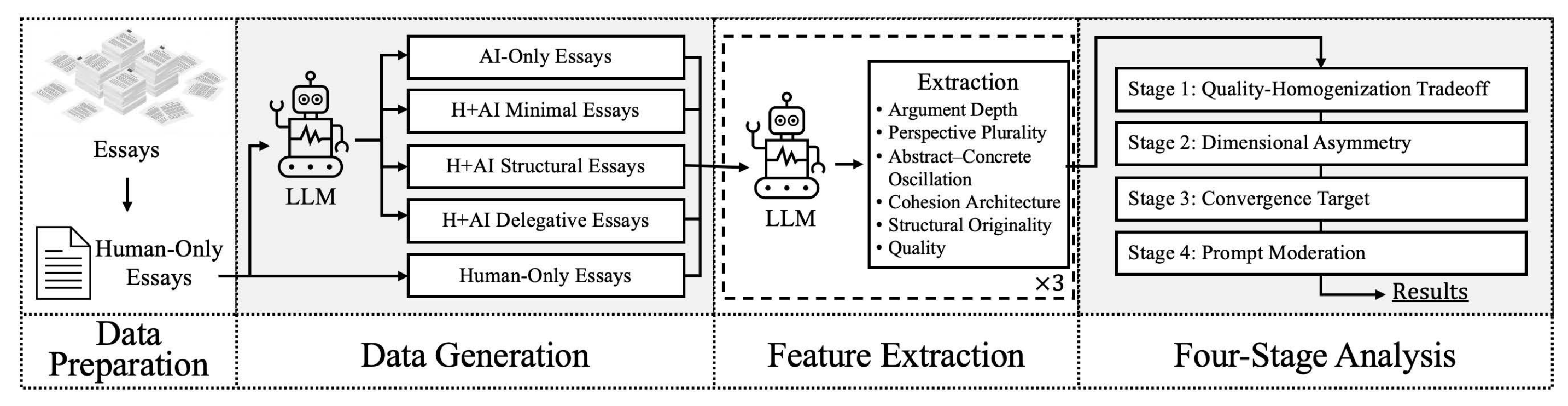}
  \caption{Overview of the experimental pipeline.}
  \label{fig:framework}
\end{figure}

The experimental pipeline consists of four phases, as illustrated in Figure~\ref{fig:framework}: i) data preparation and condition construction, ii) AI-augmented essay generation via the OpenAI API, iii) structural feature extraction through LLM-based evaluation, and iv) statistical analysis of distributional changes across conditions. All data generation, feature extraction, and analyses were implemented in Python with asynchronous parallel processing for API calls. The complete pipeline, including all prompts and parameters, was documented to ensure reproducibility.

\subsection{Base Dataset}
\label{subsec:dataset}

The base dataset was drawn from the AI Detection for Essays dataset (AIDE) corpus~\citep{31}, a publicly available dataset released through a Kaggle competition. The dataset comprises 1,378 essays, each labeled as human-written or AI-generated, spanning two essay topics: Topic~0, ``Car-free cities'' (an explanatory essay on the advantages of limiting car usage), and Topic~1, ``Does the electoral college work?'' (an argumentative essay on the U.S.\ Electoral College system). Standardized writing instructions and source materials were provided to all writers for each topic. Available metadata include character count, word count, sentence count, paragraph count, average word length, and average sentence length; rubric-based quality scores were not included.

Of the 1,378 essays, 1,375 were human-written and 3 were AI-generated. The three existing AI-generated essays were excluded from analysis due to unknown model provenance, and a matched set of AI-only essays was generated using the same model employed for the H+AI conditions.

The AIDE dataset was selected for three reasons. First, the standardized topics and instructions control for thematic variation, isolating the effect of AI augmentation on structural features. Second, the public availability of the dataset ensures reproducibility. Third, the inclusion of two distinct genres, explanatory and argumentative, enables examination of topic-dependent effects.

\subsection{Condition Construction}
\label{subsec:conditions}

Five experimental conditions were constructed, yielding a total of 6,875 essays, as summarized in Table~\ref{tab:conditions}.

\begin{table}[t]
\centering
\caption{Experimental conditions and sample sizes.}
\label{tab:conditions}
\small
\begin{tabular}{@{}llrp{5.2cm}@{}}
\toprule
Condition & Abbreviation & $n$ & Description \\
\midrule
Human-only & H & 1,375 & Student-written essays from the AIDE dataset \\
AI-only & A & 1,375 & Generated by GPT-5-mini from topic and instructions only \\
\addlinespace
H+AI (Minimal) & H+AI\textsubscript{min} & 1,375 & Augmented with ``Please improve this essay'' \\
H+AI (Structural) & H+AI\textsubscript{str} & 1,375 & Augmented with structure-specific instructions \\
H+AI (Delegative) & H+AI\textsubscript{del} & 1,375 & Augmented with ``Rewrite in your own way'' \\
\midrule
Total & & 6,875 & \\
\bottomrule
\end{tabular}
\end{table}

\begin{enumerate}[i)]
\item Condition H: All 1,375 human-written essays from the AIDE dataset were used without modification. This condition serves as the baseline for measuring structural diversity; the variance of structural features in this condition defines the reference against which homogenization is assessed.

\item Condition A: To provide a matched AI baseline, 1,375 AI-only essays were generated using GPT-5-mini~\citep{32}, maintaining the same topic distribution as condition H (Topic~0: 707 essays; Topic~1: 668 essays). Each essay was generated by providing the model with the original topic instructions and source materials from the AIDE dataset. The system prompt was set to ``You are a student writing an essay for a class assignment,'' and the temperature parameter was left at the model's default setting. This condition serves two purposes: establishing the structural fingerprint of AI-generated text and providing the reference point for convergence target analysis.

\item Condition H+AI: Each of the 1,375 human-written essays was augmented under three prompt strategies, yielding a fully crossed within-subject design. The three prompt conditions were designed to represent qualitatively distinct modes of human--AI interaction. The exact prompts used are presented in Table~\ref{tab:prompts}.
\end{enumerate}
\begin{table}[t]
\centering
\caption{Prompt conditions for AI augmentation.}
\label{tab:prompts}
\small
\begin{tabular}{@{}p{1.8cm}p{3.8cm}p{7cm}@{}}
\toprule
Condition & System prompt & User prompt \\
\midrule
Minimal & ``You are a helpful writing assistant.'' & ``Please improve the following essay. Keep the same topic and general direction. [essay]'' \\
\addlinespace
Structural & ``You are a writing tutor specializing in argumentative essay structure.'' & ``Please improve the following essay by: 1.~Strengthening the argument structure 2.~Adding counterarguments and rebuttals 3.~Improving logical coherence between paragraphs. Maintain the student's original perspective and main arguments. [essay]'' \\
\addlinespace
Delegative & ``You are an expert essay writer.'' & ``Read the following essay and rewrite it on the same topic in your own way. You may completely restructure and rewrite the content. [essay]'' \\
\bottomrule
\end{tabular}
\end{table}

The Minimal condition reproduces the most common student use pattern: pasting an essay with a vague request for improvement. The Structural condition represents directed augmentation with explicit structural targets. The Delegative condition approximates full delegation, in which the student effectively outsources the writing task.

All generation was performed via the OpenAI API using GPT-5-mini with the max\_completion\_tokens parameter set to 2,048. To manage rate limits and reduce generation time, 10 API keys were used in a round-robin configuration with 50 concurrent requests. The generation process incorporated automatic retry logic (exponential backoff with a maximum of 5 retries) and periodic checkpoint saving. A small number of API calls failed after all retries; supplementary generation rounds using the same procedure brought all conditions to the target count of 1,375.

\subsection{Structural Feature Extraction}
\label{subsec:features}

All 6,875 essays were evaluated on six structural dimensions using GPT-5 as an automated evaluator. The scales and operational definitions of the six dimensions are presented in Table~\ref{tab:dimensions}.

\begin{table}[t]
\centering
\caption{Structural dimensions and operational definitions.}
\label{tab:dimensions}
\small
\begin{tabular}{@{}p{3.2cm}cp{7.5cm}@{}}
\toprule
Dimension & Scale & Operational definition \\
\midrule
Argument Depth & 1--5 & Number of inferential layers between claims and evidence. 1 = flat assertions without support; 5 = deep inferential chains. \\
\addlinespace
Perspective Plurality & 1--5 & Number and integration quality of distinct viewpoints. 1 = single perspective; 5 = multiple perspectives with synthesis and evaluation. \\
\addlinespace
Abstract--Concrete Oscillation & 1--5 & Frequency and purposefulness of transitions between abstract claims and concrete examples. \\
\addlinespace
Cohesion Architecture & 1--5 & Quality of logical connections between and within paragraphs. 1 = fragmented; 5 = dense logical connectives with clear progression. \\
\addlinespace
Structural Originality & 1--5 & Degree of deviation from standard organizational templates. \\
\addlinespace
Quality (Holistic) & 1--6 & Overall essay quality. 1 = very poor; 6 = excellent. \\
\bottomrule
\end{tabular}
\end{table}

The extraction procedure was as follows. GPT-5 was provided with a structured prompt containing anchored descriptions for each scale point across all six dimensions and was instructed to return only a valid JSON object containing the six numeric scores. The response format was constrained to JSON output. The max\_completion\_tokens parameter was set to 4,096 to accommodate the model's internal reasoning process. Each essay was evaluated three times independently, and the mean of the three scores was used for analysis. This triple-extraction design enables assessment of measurement stability.

Extraction was parallelized using asynchronous API calls with 10 API keys and 50 concurrent requests. Across 6,875 essays $\times$ 3 runs = 20,625 API calls, zero missing values were produced. All essays yielded valid scores on all six dimensions across all three runs.

\subsection{Validation of Feature Extraction}
\label{subsec:validation}

The reliability of LLM-based structural feature extraction was assessed through two validation procedures prior to the main analysis. First, regarding extraction stability, the Coefficient of Variation ($\mathrm{CV} = \sigma / \mu$) across the three independent extraction runs was computed for each essay on each dimension. A threshold of $\mathrm{CV} < 0.10$ was adopted as the stability criterion. All six dimensions met this criterion at the population level: Argument Depth (mean $\mathrm{CV} = 0.041$), Perspective Plurality ($0.070$), Abstract--Concrete Oscillation ($0.046$), Cohesion Architecture ($0.031$), Structural Originality ($0.065$), and Quality ($0.029$). Individual-level maximum CVs ranged from $0.354$ to $0.408$, indicating that a small number of structurally ambiguous essays produced greater variability in evaluation. The use of mean scores across three runs mitigates this variability.

Second, regarding topic bias, Mann--Whitney $U$ tests were conducted on condition H essays to assess whether structural features differed systematically between the two topics. Significant differences ($p < 0.05$) were detected for five of six dimensions; only Argument Depth showed no significant difference ($p = 0.413$). This is an expected result given the different genres and does not threaten the validity of within-condition variance comparisons.

\subsection{Analytical Framework}
\label{subsec:analysis}

The analysis proceeds in four stages, each corresponding to a research question.

Stage~1: Quality-Homogenization Tradeoff. For each H+AI condition, the mean quality score is compared with condition H using Welch's $t$-test (effect size: Cohen's $d$). Simultaneously, for each structural dimension, the variance is compared between condition H and each H+AI condition using the Brown--Forsythe test (a robust variant of Levene's test using medians). The variance ratio is defined as:
\begin{equation}
\mathrm{VR} = \frac{\sigma^2_{\mathrm{H}}}{\sigma^2_{\mathrm{H+AI}}}
\label{eq:vr}
\end{equation}
$\mathrm{VR} > 1$ indicates homogenization (variance reduction). Bootstrap confidence intervals (10,000 resamples) are computed for each VR. The Quality-Homogenization Tradeoff is operationalized as the co-occurrence of significant quality improvement ($p < 0.05$) and significant variance reduction ($\mathrm{VR} > 1$, $p < 0.05$) on at least one structural dimension.

Stage~2: Dimensional Asymmetry. For each dimension $d$ and each H+AI condition, the Homogenization Index (HI) is computed:
\begin{equation}
\mathrm{HI}_d = 1 - \frac{\sigma^2_{\mathrm{H+AI},d}}{\sigma^2_{\mathrm{H},d}}
\label{eq:hi}
\end{equation}
$\mathrm{HI} > 0$ indicates homogenization, $\mathrm{HI} < 0$ indicates diversification, and $\mathrm{HI} = 0$ indicates no change. The five-dimensional HI vector constitutes the homogenization profile of each prompt condition.

Stage~3: Convergence Target. The centroid of each condition is computed in the five-dimensional structural space (excluding Quality). All dimensions are $z$-score standardized prior to analysis. The Replacement Ratio (RR) is defined as:
\begin{equation}
\mathrm{RR} = \frac{d(\mathrm{H+AI},\;\mathrm{H})}{d(\mathrm{H+AI},\;\mathrm{H}) + d(\mathrm{H+AI},\;\mathrm{A})}
\label{eq:rr}
\end{equation}
where $d$ denotes Euclidean distance. $\mathrm{RR} > 0.5$ indicates that H+AI is closer to A than to H. To test for the emergence of a structurally novel region, the perpendicular distance from the H+AI centroid to the H--A axis is computed, and its significance is assessed via permutation testing (10,000 permutations). The null hypothesis is that H+AI lies on the H--A line segment. The full five-condition structure is visualized using UMAP ($n\_\mathrm{neighbors} = 15$, $\mathrm{min\_dist} = 0.1$) for dimensionality reduction.

Stage~4: Prompt Moderation. The three H+AI conditions are compared on all metrics from Stages~1--3. Kruskal--Wallis tests assess whether mean structural feature values differ across prompt conditions, and Levene's tests assess whether variances differ. A significant interaction between prompt type and structural dimension, that is, a pattern in which the prompt effect on homogenization differs by dimension, would demonstrate that homogenization is not an intrinsic property of AI augmentation but a function of interaction design.

All statistical tests use $\alpha = 0.05$ with Bonferroni correction where applicable. Analyses were conducted using Python~3 with SciPy, NumPy, pandas, and UMAP-learn.

%% file: Authors/documents/4_experiment-and-analysis.tex

\section{Results and Analysis}
\label{sec:results}

\subsection{Dataset and Descriptive Statistics}
\label{subsec:descriptive}

The final dataset comprises 6,875 essays across five conditions, with 1,375 essays per condition. The fully crossed design ensures that each of the 1,375 human essays has a corresponding augmented version in each of the three H+AI conditions, enabling paired comparisons.

Table~\ref{tab:textchars} presents text-level characteristics by condition. AI augmentation produced notable differences in essay length. The Structural condition generated the longest texts (mean = 5,111 characters, exceeding even the AI-only mean of 4,538), whereas the Minimal and Delegative conditions produced texts shorter than the human originals. This length pattern suggests that different prompt strategies elicit qualitatively distinct AI responses.

\begin{table}[t]
\centering
\caption{Text-level characteristics by condition.}
\label{tab:textchars}
\small
\begin{tabular}{@{}lrr@{}}
\toprule
Condition & Mean characters (SD) & Difference from H \\
\midrule
H & 3,173 (918) & --- \\
A & 4,538 (533) & +1,365 \\
H+AI Minimal & 2,865 (600) & $-$308 \\
H+AI Structural & 5,111 (720) & +1,938 \\
H+AI Delegative & 2,887 (588) & $-$286 \\
\bottomrule
\end{tabular}
\end{table}

Table~\ref{tab:descriptive} presents the scores on all six structural dimensions by condition. Mean values on all structural features shifted upward from condition H to the AI-involved conditions. Two patterns in the standard deviations, the focus of this study, are particularly noteworthy. First, the standard deviation of Cohesion Architecture was compressed from 0.47 in condition H to 0.22--0.26 across all H+AI conditions, suggesting strong homogenization on this dimension at the descriptive level. Second, on Argument Depth, only the Structural condition exhibited a standard deviation (0.37) larger than condition H (0.27), while the Minimal and Delegative conditions showed reduced values (0.19--0.20), foreshadowing the prompt-dependent divergence in homogenization patterns examined in subsequent stages.

\begin{table}[t]
\centering
\caption{Structural feature scores by condition: Mean (SD).}
\label{tab:descriptive}
\small
\begin{tabular}{@{}lccccc@{}}
\toprule
Dimension & H & A & Minimal & Structural & Delegative \\
\midrule
Argument Depth & 2.13 (0.27) & 3.37 (0.35) & 3.04 (0.20) & 3.36 (0.37) & 3.05 (0.19) \\
Perspective Plurality & 1.62 (0.51) & 2.51 (0.58) & 1.97 (0.58) & 3.43 (0.52) & 2.07 (0.62) \\
Abstract--Concrete Osc. & 2.84 (0.39) & 3.76 (0.35) & 3.37 (0.45) & 3.53 (0.46) & 3.37 (0.44) \\
Cohesion Architecture & 2.64 (0.47) & 4.11 (0.30) & 4.09 (0.22) & 3.99 (0.26) & 4.15 (0.26) \\
Structural Originality & 1.51 (0.49) & 2.01 (0.11) & 1.74 (0.39) & 2.28 (0.41) & 1.87 (0.34) \\
Quality & 2.69 (0.47) & 4.34 (0.41) & 4.55 (0.44) & 4.23 (0.35) & 4.74 (0.38) \\
\bottomrule
\end{tabular}
\end{table}

All three H+AI conditions exhibited significant quality improvements over condition H, as shown in Table~\ref{tab:quality}. Effect sizes were uniformly large (Cohen's $d > 3.7$, all $p < 0.0001$), indicating that AI augmentation substantially improved essay quality regardless of prompt strategy. The Delegative condition showed the largest improvement ($d = 4.814$), followed by Minimal ($d = 4.102$) and Structural ($d = 3.754$).

\begin{table}[t]
\centering
\caption{Quality improvement by condition.}
\label{tab:quality}
\small
\begin{tabular}{@{}lcrc@{}}
\toprule
Condition & H mean $\to$ H+AI mean & Cohen's $d$ & $p$ \\
\midrule
H+AI Minimal & 2.69 $\to$ 4.55 & 4.102 & $< 0.0001$ \\
H+AI Structural & 2.69 $\to$ 4.23 & 3.754 & $< 0.0001$ \\
H+AI Delegative & 2.69 $\to$ 4.74 & 4.814 & $< 0.0001$ \\
\bottomrule
\end{tabular}
\end{table}

Alongside these quality gains, significant variance reductions were detected on specific structural dimensions, as shown in Table~\ref{tab:vr}. The variance ratio $\mathrm{VR} = \sigma^2_{\mathrm{H}} / \sigma^2_{\mathrm{H+AI}}$ was computed for each dimension and condition, with 95\% bootstrap confidence intervals (10,000 resamples). Cases in which $\mathrm{VR} > 1$ with $p < 0.05$ (Brown--Forsythe test) were classified as statistically significant homogenization.

\begin{table}[t]
\centering
\caption{Variance ratio by dimension and prompt condition.}
\label{tab:vr}
\small
\begin{tabular}{@{}lccc@{}}
\toprule
Dimension & Minimal VR [95\% CI] & Structural VR [95\% CI] & Delegative VR [95\% CI] \\
\midrule
Argument Depth & 1.881 [1.52, 2.36] & 0.528 [0.46, 0.60] & 2.065 [1.66, 2.64] \\
Perspective Plurality & 0.771 [0.70, 0.85] & 0.959 [0.87, 1.07] & 0.685 [0.62, 0.76] \\
Abstract--Concrete Osc. & 0.737 [0.66, 0.82] & 0.711 [0.63, 0.80] & 0.761 [0.68, 0.85] \\
Cohesion Architecture & 4.554 [3.88, 5.45] & 3.160 [2.72, 3.73] & 3.180 [2.81, 3.62] \\
Structural Originality & 1.563 [1.41, 1.73] & 1.413 [1.27, 1.57] & 2.082 [1.84, 2.36] \\
\bottomrule
\end{tabular}
\end{table}

\begin{figure}[t]
  \centering
  \includegraphics[width=0.85\textwidth]{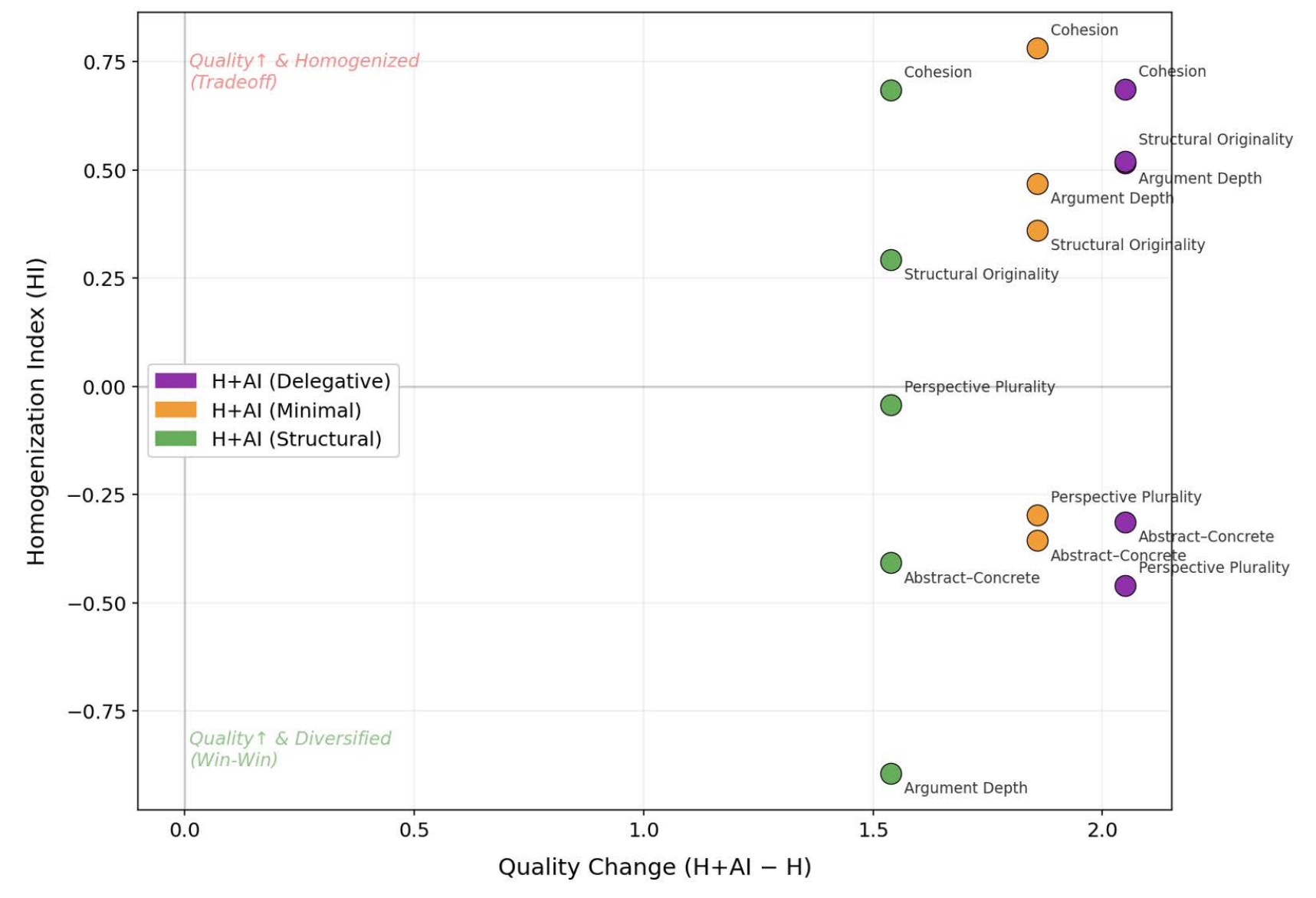}
  \caption{Quality-Homogenization Tradeoff. }
  \label{fig:tradeoff}
\end{figure}

The Quality-Homogenization Tradeoff was confirmed across all three conditions. Quality improved significantly while homogenization was detected on at least two structural dimensions. The Minimal and Delegative conditions showed homogenization on three dimensions (Argument Depth, Cohesion Architecture, Structural Originality), whereas the Structural condition showed homogenization on only two (Cohesion Architecture, Structural Originality). Notably, Perspective Plurality and Abstract--Concrete Oscillation exhibited $\mathrm{VR} < 1$ across all conditions, indicating diversification rather than homogenization. The relationship between quality change and homogenization is visualized in Figure~\ref{fig:tradeoff}, which plots quality change (H+AI $-$ H) on the horizontal axis and the HI on the vertical axis for 3 conditions $\times$ 5 dimensions = 15 data points. All points fall in the positive region of the horizontal axis (quality improvement), while the vertical axis clearly separates Cohesion Architecture and Structural Originality in the upper half (homogenization) from Perspective Plurality and Abstract--Concrete Oscillation in the lower half, visually confirming that quality improvement is accompanied by both homogenization and diversification.

\subsection{Evaluation of Dimensional Asymmetry}
\label{subsec:rq2}

The Homogenization Index ($\mathrm{HI} = 1 - \sigma^2_{\mathrm{H+AI}} / \sigma^2_{\mathrm{H}}$) was computed for each dimension and condition, as shown in Table~\ref{tab:hi}. HI values revealed pronounced asymmetry across the five dimensions. Cohesion Architecture recorded the highest HI values across all three conditions ($+0.68$ to $+0.78$), indicating that 68--78\% of the original variance was eliminated by AI augmentation. This reflects a strong, prompt-independent standardization of inter-paragraph logical connectivity. Structural Originality also showed consistently positive HI values ($+0.29$ to $+0.52$), confirming that AI tends to normalize organizational patterns toward standard templates.
\begin{table}[t]
\centering
\caption{Homogenization Index by dimension and prompt condition.}
\label{tab:hi}
\small
\begin{tabular}{@{}lrrrrr@{}}
\toprule
Dimension & Minimal & Structural & Delegative & Mean & Direction \\
\midrule
Cohesion Architecture & +0.780 & +0.684 & +0.686 & +0.717 & Strong homog. \\
Structural Originality & +0.360 & +0.292 & +0.520 & +0.391 & Moderate homog. \\
Argument Depth & +0.468 & $-$0.894 & +0.516 & +0.030 & Prompt-dependent \\
Perspective Plurality & $-$0.297 & $-$0.043 & $-$0.461 & $-$0.267 & Diversification \\
Abstract--Concrete Osc. & $-$0.357 & $-$0.407 & $-$0.314 & $-$0.359 & Diversification \\
\bottomrule
\end{tabular}
\end{table}

In contrast, Perspective Plurality ($\mathrm{HI} = -0.04$ to $-0.46$) and Abstract--Concrete Oscillation ($\mathrm{HI} = -0.31$ to $-0.41$) exhibited negative HI values across all conditions. On these dimensions, inter-student variability increased after AI augmentation, a change opposite to homogenization.

The most noteworthy pattern is Argument Depth. This dimension showed homogenization under the Minimal ($\mathrm{HI} = +0.47$) and Delegative ($\mathrm{HI} = +0.52$) conditions but strong diversification under the Structural condition ($\mathrm{HI} = -0.89$). The mean HI across the three conditions is approximately zero ($+0.03$), but this does not indicate the absence of an effect; rather, it reflects the cancellation of opposing effects across prompt conditions. This reversal is examined in detail in Section~\ref{subsec:rq4}.

These results demonstrate that AI augmentation does not uniformly homogenize student writing. Rather, the effect is dimension-specific: Cohesion Architecture and Structural Originality are systematically homogenized, Abstract--Concrete Oscillation and Perspective Plurality are diversified, and Argument Depth depends on the prompt strategy employed.

\subsection{Convergence Target Analysis}
\label{subsec:rq3}

To identify where homogenized essays converge in structural space, centroids for all five conditions were computed in the $z$-score standardized five-dimensional feature space (excluding Quality). Table~\ref{tab:convergence} reports inter-centroid distances and the RR.

\begin{table}[t]
\centering
\caption{Inter-centroid distances and convergence indicators.}
\label{tab:convergence}
\small
\begin{tabular}{@{}lrrrrc@{}}
\toprule
Condition & $d$(H+AI, H) & $d$(H+AI, A) & RR & Perp.\ dist. & Emergence $p$ \\
\midrule
H+AI Minimal & 3.035 & 1.302 & 0.700 & 0.781 & $< 0.001$ \\
H+AI Structural & 4.339 & 1.345 & 0.763 & 1.336 & $< 0.001$ \\
H+AI Delegative & 3.189 & 1.136 & 0.737 & 0.731 & $< 0.001$ \\
\bottomrule
\end{tabular}
\end{table}

All three conditions yielded $\mathrm{RR} > 0.7$, indicating that H+AI essays are structurally far closer to AI-only essays than to human-only essays. This suggests that, regardless of prompt type, AI augmentation pulls student writing toward AI structural patterns.

However, the permutation test for emergence (10,000 permutations) was simultaneously significant at $p < 0.001$ for all conditions, indicating that H+AI centroids deviated significantly from the H--A axis. That is, H+AI essays do not simply occupy a point on the line between H and A but form a structurally distinct region.

\begin{figure}[t]
  \centering
  \includegraphics[width=0.85\textwidth]{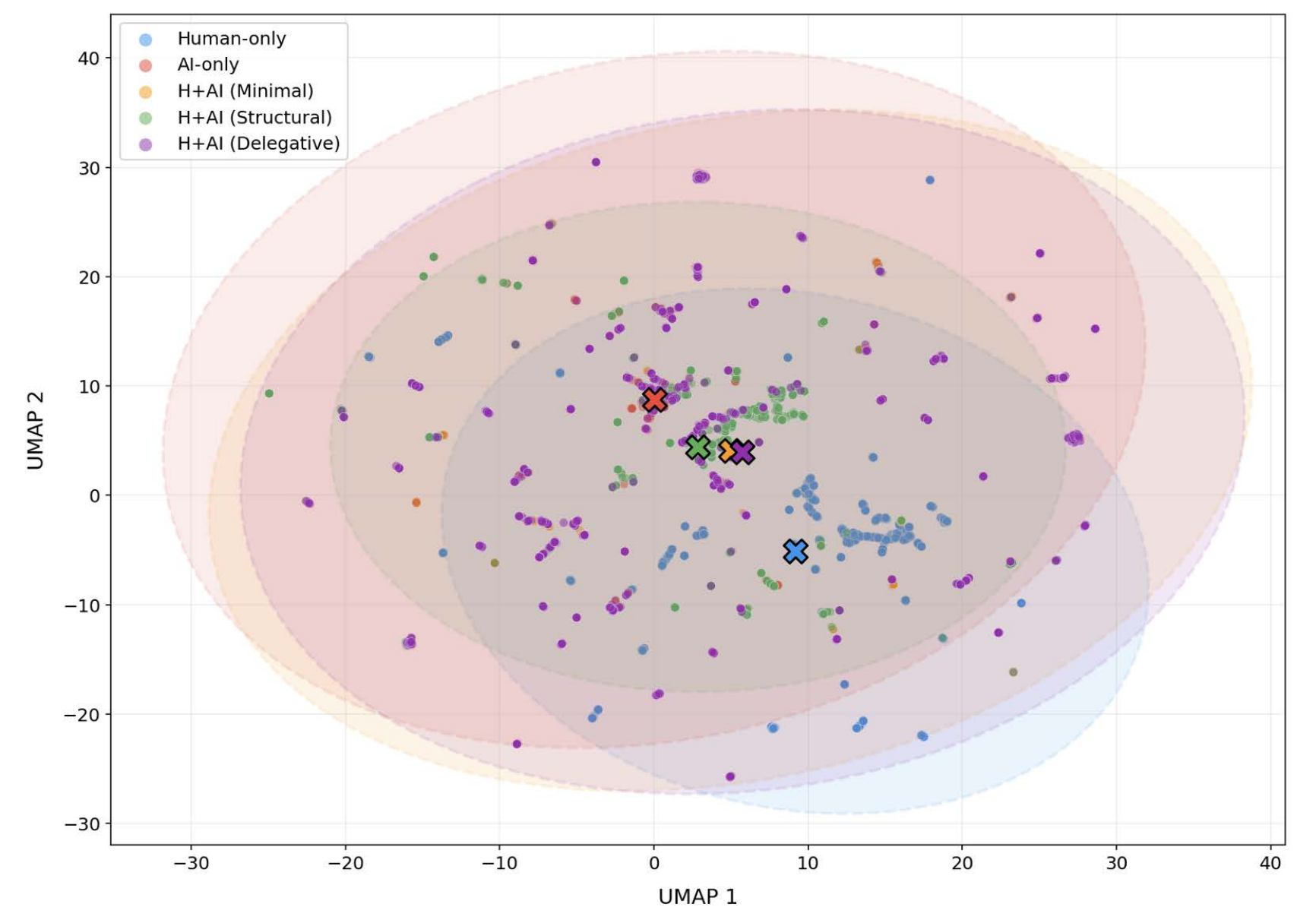}
  \caption{UMAP projection of the five-dimensional structural space. }
  \label{fig:umap}
\end{figure}

Figure~\ref{fig:umap} presents the UMAP projection of the five-condition structural space. Condition H points (blue) are distributed with broad spread toward the lower-right region, while condition A points (red) cluster more tightly toward the upper-left. The larger 95\% confidence ellipse for condition H relative to condition A reflects the greater structural diversity of human essays compared to AI-generated essays. The centroids ($\times$ markers) of the three H+AI conditions are positioned near the condition A centroid but do not coincide with it; notably, the Structural condition centroid (green) is located farthest from the H--A axis. This spatial configuration visually corroborates that H+AI essays are pulled toward AI structural patterns while deviating into a distinct region of the structural space.

The simultaneous detection of replacement ($\mathrm{RR} > 0.7$) and emergence (significant perpendicular distance) reveals a composite phenomenon. AI augmentation strongly pulls student writing toward AI structural patterns, yet the resulting essays are not simple copies of AI output. Human input, filtered through the interaction process, gives rise to hybrid structural profiles that occupy a novel region of the structural space.

\subsection{Prompt Moderation Analysis}
\label{subsec:rq4}

Kruskal--Wallis tests confirmed that all five structural dimensions differed significantly across the three prompt conditions (all $p < 0.0001$). Levene's tests indicated that variances also differed significantly across prompt conditions for four of five dimensions, as shown in Table~\ref{tab:moderation}.
\begin{table}[t]
\centering
\caption{Prompt moderation effects: between-condition comparisons.}
\label{tab:moderation}
\small
\begin{tabular}{@{}lrrcc@{}}
\toprule
Dimension & Levene's $p$ & KW $p$ & Variance diff. & Mean diff. \\
\midrule
Argument Depth & $< 0.001$ & $< 0.001$ & Yes & Yes \\
Perspective Plurality & 0.006 & $< 0.001$ & Yes & Yes \\
Abstract--Concrete Osc. & 0.073 & $< 0.001$ & No & Yes \\
Cohesion Architecture & $< 0.001$ & $< 0.001$ & Yes & Yes \\
Structural Originality & $< 0.001$ & $< 0.001$ & Yes & Yes \\
\bottomrule
\end{tabular}
\end{table}

\begin{figure}[t]
  \centering
  \includegraphics[width=0.75\textwidth]{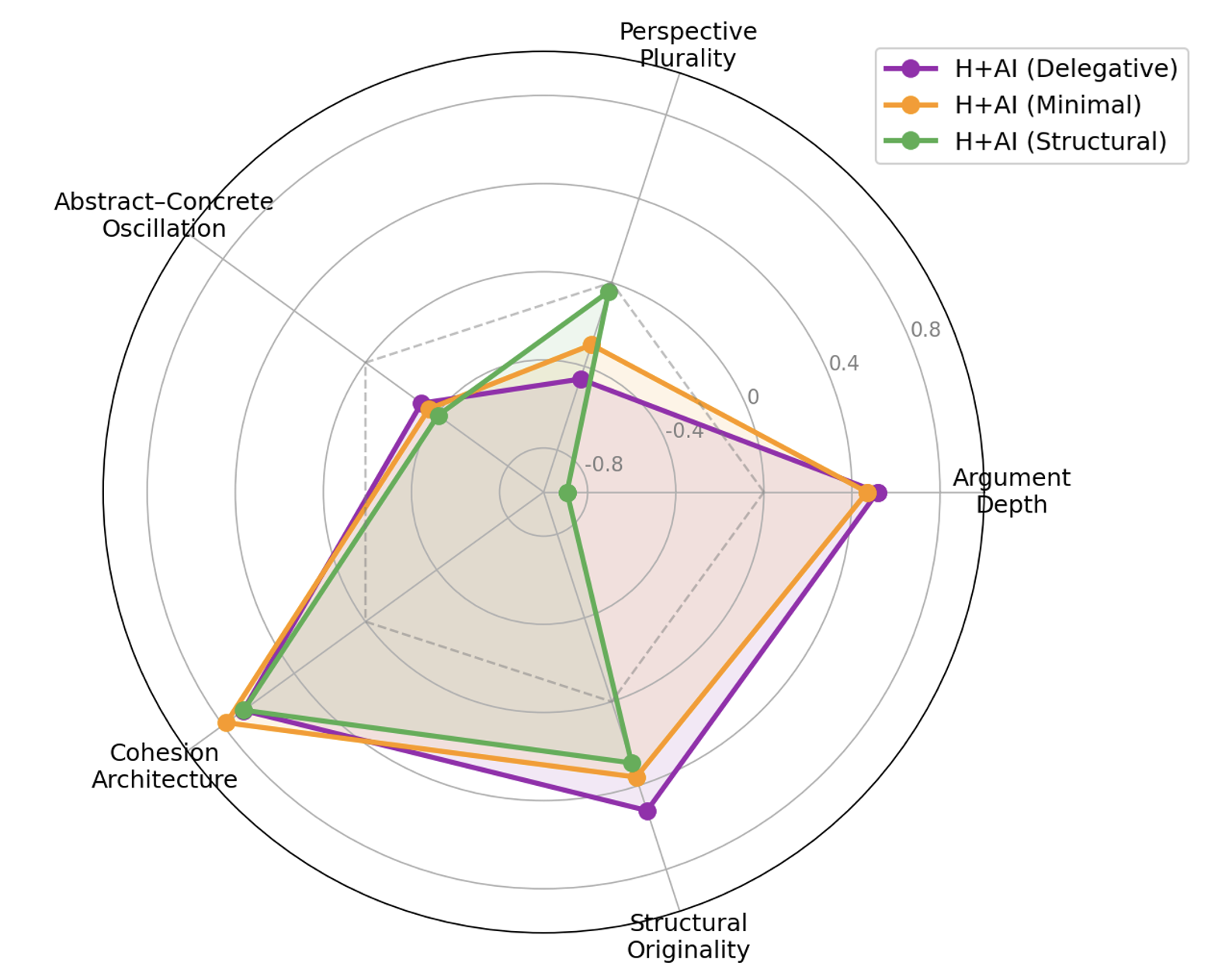}
  \caption{Radar plot of homogenization profiles across three prompt conditions. }
  \label{fig:radar}
\end{figure}

The most important finding is the reversal of homogenization on Argument Depth. The Minimal condition ($\mathrm{HI} = +0.47$) and the Delegative condition ($\mathrm{HI} = +0.52$) homogenized Argument Depth, while the Structural condition ($\mathrm{HI} = -0.90$) produced dramatic diversification. This means that even when the same AI model is applied to the same input essays, prompt differences alone can produce either homogenization or diversification. This reversal is the most visually salient feature in the radar plot, as shown in Figure~\ref{fig:radar}: the Structural condition line (green) swings sharply in the opposite direction from the other two conditions on the Argument Depth axis, making it immediately apparent that the same AI model can exert diametrically opposite effects on structural diversity depending on the prompt. This figure constitutes the most direct visual evidence that homogenization is not an intrinsic property of AI but a function of interaction design.

\subsection{Qualitative Analysis}
\label{subsec:qualitative}

To ground the statistical findings at the text level, specific essays were examined across conditions.

Condition H essays were divided into terciles by cohesion score, and their post-augmentation values under the Minimal condition were tracked, as presented in Table~\ref{tab:convergence_cohesion}. The original range of 2.09--3.49 (a spread of 1.40) collapsed to 4.07--4.12 after augmentation (a spread of only 0.05). The same pattern was confirmed under the Structural and Delegative conditions.

\begin{table}[t]
\centering
\caption{Convergence pattern of Cohesion Architecture by original level (Minimal condition).}
\label{tab:convergence_cohesion}
\small
\begin{tabular}{@{}lrrr@{}}
\toprule
Original level & Original mean & Post-augmentation mean & Change \\
\midrule
Low tercile & 2.09 & 4.07 & +1.98 \\
Middle tercile & 2.92 & 4.09 & +1.17 \\
High tercile & 3.49 & 4.12 & +0.63 \\
\bottomrule
\end{tabular}
\end{table}

At the text level, low-cohesion essays characterized by grammatical errors and fragmented paragraph structures were transformed into well-organized texts with explicit connectives and clear paragraph progression. In contrast, essays that already had high cohesion remained unchanged at 4.0 after augmentation. This pattern suggests that AI possesses an internal target level for cohesion of approximately 4.0 and pulls all essays toward that level regardless of their starting point.

A similar bidirectional convergence was observed for Structural Originality. Essays with high originality scores (score $\geq$ 2.5, $n = 34$) declined from a mean of 2.86 to 2.05 ($-0.81$), while essays with low originality scores (score $\leq$ 1.0, $n = 505$) rose from a mean of 1.00 to 1.72 ($+0.72$). Both groups converged to approximately 1.7--2.0, reflecting the same target-level attractor mechanism observed for Cohesion Architecture and indicating that AI normalizes organizational patterns toward standard templates.

A single essay (original Argument Depth = 2.0) was tracked under both Minimal and Structural augmentation. Under the Minimal condition, depth increased slightly to 2.7; AI polished the prose but preserved the original argumentative skeleton. Under the Structural condition, depth rose substantially to 4.0, with the AI introducing a clear thesis statement, multi-layered reasoning, counterarguments, and rebuttals. This illustrates the mechanism underlying the HI reversal: vague prompts produce uniform, shallow improvement, whereas specific structural instructions elicit input-dependent, differentiated deepening.

\subsection{Topic-Level Robustness Check}
\label{subsec:robustness}

Given the topic effects identified in the validation analysis (Section~\ref{subsec:validation}), all primary analyses were repeated separately for Topic~0 (Car-free cities, explanatory essays, $n = 3{,}535$) and Topic~1 (Does the electoral college work?, argumentative essays, $n = 3{,}340$). Table~\ref{tab:robustness} summarizes the consistency of findings across topics.

\begin{table}[t]
\centering
\caption{Consistency of main findings across topics.}
\label{tab:robustness}
\small
\begin{tabular}{@{}p{4cm}p{3.5cm}p{3.5cm}c@{}}
\toprule
Finding & Topic 0 & Topic 1 & Consistency \\
\midrule
Quality improvement & $d = 3.71$--$5.49$ & $d = 4.07$--$4.59$ & Consistent \\
Cohesion homog. & HI $= +0.59$ to $+0.81$ & HI $= +0.64$ to $+0.79$ & Consistent \\
Struct.\ Orig.\ homog. & HI $= +0.14$ to $+0.42$ & HI $= +0.47$ to $+0.63$ & Consistent \\
Arg.\ Depth reversal & HI $= -0.96$ (Str) & HI $= -0.64$ (Str) & Consistent \\
Abstract--Concrete Osc. & Diversified & Homogenized & Reversed \\
Perspective Plurality & $\approx 0$ & Diversified & Partial \\
\bottomrule
\end{tabular}
\end{table}

Three of the four main findings, the Quality-Homogenization Tradeoff, consistent homogenization of Cohesion Architecture and Structural Originality, and the prompt-dependent reversal on Argument Depth, replicated fully across both topics. The reversal on Argument Depth, the study's most novel finding, was robust: the Structural condition produced strong diversification on both the explanatory topic ($\mathrm{HI} = -0.96$) and the argumentative topic ($\mathrm{HI} = -0.64$), while Minimal and Delegative produced homogenization on both.

Two dimensions exhibited topic-dependent effects. Abstract--Concrete Oscillation was strongly diversified on the explanatory topic but mildly homogenized on the argumentative topic. Perspective Plurality was approximately zero on the explanatory topic but diversified on the argumentative topic. These genre-dependent effects suggest that the impact of AI on specific structural dimensions is moderated by the rhetorical demands of the writing task, with implications for the cross-genre generalizability of the Homogenization Hypothesis.

%% file: Authors/documents/5_discussion.tex

\section{Discussion}
\label{sec:discussion}

This study examined whether AI augmentation improves student writing quality while simultaneously homogenizing its structural patterns, across 6,875 essays, six structural dimensions, and three prompt conditions. This section organizes the novel findings, discusses their theoretical and practical implications, and addresses limitations.

\subsection{Key Findings}
\label{subsec:findings}

The results update the current understanding of AI and student writing in several respects. First, quality improvement and structural homogenization co-occur. Previous research, which evaluated AI augmentation solely on the quality dimension, was structurally incapable of detecting this phenomenon. By independently measuring quality and structural diversity, this study demonstrated for the first time that Cohen's $d$ of 3.7--4.8 in quality improvement co-occurs with up to 4.6-fold variance compression in Cohesion Architecture within the same dataset. This result suggests that conventional quality-only evaluation may have been overlooking homogenization as a significant side effect of AI augmentation.

Second, homogenization is dimension-specific, not uniform. Both the claim that AI homogenizes thinking and the claim that AI diversifies thinking are empirically inaccurate. The reality is dimension-dependent. Cohesion Architecture ($\mathrm{HI} = +0.68$ to $+0.78$) and Structural Originality ($\mathrm{HI} = +0.29$ to $+0.52$) were consistently homogenized across all conditions, whereas Perspective Plurality and Abstract--Concrete Oscillation were diversified. This asymmetry indicates that AI does not exert the same effect on all structural features. AI tends to strongly standardize lower-order structural features such as paragraph connectivity patterns and template-like organizational structures, while exerting weaker standardization pressure on higher-order cognitive features such as multi-perspective synthesis and abstract-concrete reasoning.

Third, prompt design can reverse homogenization into diversification, even with the same AI model. This is the study's most novel finding. On the single dimension of Argument Depth, the Minimal condition ($\mathrm{HI} = +0.47$) and the Delegative condition ($\mathrm{HI} = +0.52$) produced homogenization, while the Structural condition ($\mathrm{HI} = -0.89$) produced substantial diversification. This reversal replicated across both topics. The implication is that homogenization is not an intrinsic property of AI but a design variable of human--AI interaction. A vague instruction such as ``improve this essay'' cedes full control over argumentative structure to the AI, allowing convergence toward its internal optimal template. In contrast, a specific instruction such as ``strengthen the argument structure and add counterarguments'' forces the AI to respond in an input-dependent manner, producing differentiated deepening that varies across students and thereby resulting in diversification.

Fourth, AI-augmented essays form a distinct third structural space. In the convergence target analysis (RQ3), H+AI essays were strongly pulled toward AI structural patterns ($\mathrm{RR} > 0.7$) across all three conditions while simultaneously deviating significantly from the H--A axis (all $p < 0.001$). This indicates that H+AI essays are not simple copies of AI output but possess a structurally distinct profile arising from the non-linear combination of human input and AI processing. The Structural condition exhibited the largest deviation from the H--A axis (perpendicular distance = 1.336), suggesting that specific structural instructions promote the formation of the most distinctive structural space.

Fifth, the mechanism of homogenization is gravitational pull toward a target level. Qualitative analysis revealed that AI possesses internal target levels for each structural dimension (Cohesion Architecture $\approx$ 4.0, Argument Depth $\approx$ 3.0, Structural Originality $\approx$ 2.0) and pulls all essays toward these targets regardless of starting point. For Cohesion Architecture, the low tercile (original 2.09) was raised to 4.07, the middle tercile (2.92) to 4.09, and the high tercile (3.49) to 4.12. An original inter-group spread of 1.40 collapsed to just 0.05. This attractor model explains why quality improvement and homogenization are two sides of the same mechanism.

\subsection{Theoretical Implications}
\label{subsec:theoretical}

The concept of cognitive boundary permeability introduced in Section~\ref{subsec:what_changed}, the claim that generative AI alters permeability at the process layer of cognition rather than merely the input layer, received empirical support from this study. Structural features such as Cohesion Architecture and Structural Originality are products of reasoning and compositional processes. Their systematic alteration by AI, manifesting as gravitational pull toward internal target levels, constitutes direct evidence that AI intervenes in the structuring and argumentation processes themselves, not merely in information provision. The simultaneous finding that Perspective Plurality and Abstract--Concrete Oscillation resist homogenization suggests that process-layer permeability is itself dimension-dependent. AI penetrates deeply into formal processes such as paragraph connectivity and organizational structure but shallowly into content-dependent processes such as multi-perspective synthesis and abstract-concrete reasoning. This leads to a more refined understanding: cognitive boundary permeability is not uniform but varies by the type of cognitive process involved.

The concept of the ACU proposed in Section~\ref{subsec:what_changed} was supported by the convergence target analysis. The finding that H+AI essays formed a structurally distinct region deviating significantly from the H--A axis across all three conditions provides evidence that ACU possesses emergent properties different from both human and AI structures. Furthermore, the degree of this deviation varied by prompt condition, demonstrating that ACU properties are strongly dependent on the Interaction term. ACU is not a fixed entity but a dynamic system whose structural characteristics vary with interaction design. Under vague prompts, ACU output is dominated by AI patterns (partial replacement); under specific structural instructions, ACU forms a distinctive structural space (partial emergence).

The Homogenization Hypothesis was partially supported. AI augmentation improves quality while simultaneously inducing homogenization on specific dimensions, but the effect is not uniform; homogenization, no change, and diversification emerge on different dimensions in different patterns. This result calls for refining the original hypothesis into a Dimension-Specific Homogenization Hypothesis: the homogenizing effect of AI augmentation depends on the structural dimension in question, with lower-order structural features being more susceptible and higher-order cognitive features being more resistant.

Finally, among the three convergence scenarios proposed in the Introduction, neither pure replacement nor pure blending alone accounts for the data. The simultaneous detection of replacement and emergence across all conditions indicates that AI augmentation is a composite phenomenon, partial replacement with partial emergence, in which H+AI essays are pulled toward AI structural patterns while human input contributes not residually but structurally, occupying a distinct position off the H--A axis. This is consistent with the ACU framework, in which the output is neither simple addition (H + A) nor simple replacement ($\to$ A) but a product of non-linear combination.

\subsection{Practical Implications}
\label{subsec:practical}

The results provide concrete, actionable implications for educational practice. First, AI use policies should be designed on a dimension-by-dimension basis. For assignments targeting cohesion improvement (e.g., practicing logical paragraph connectivity), permitting AI use entails limited educational loss from homogenization, because cohesion improvement is a domain where AI excels and the direction of improvement is generally educationally desirable. In contrast, for assignments that emphasize structural originality (e.g., constructing an argument from a unique perspective), AI use poses a high risk of eroding the diversity of students' organizational patterns, and restriction is warranted. The blanket policies, either full prohibition or full permission of AI, currently adopted by many institutions ignore this dimension-dependence and simultaneously impose excessive restriction on some dimensions and insufficient protection on others.

Second, prompt literacy education is an urgent priority. The most practically important finding of this study is that prompt specificity controls homogenization. A vague instruction such as ``improve this essay'' maximizes cohesion homogenization ($\mathrm{HI} = +0.78$) and also homogenizes Argument Depth ($\mathrm{HI} = +0.47$). A specific structural instruction such as ``strengthen the argument structure'' cannot avoid cohesion homogenization ($\mathrm{HI} = +0.68$) but produces substantial diversification on Argument Depth ($\mathrm{HI} = -0.89$). This means that how students interact with AI fundamentally determines the educational consequences of AI use. In writing education, teaching prompt design is as important as teaching writing itself.

Third, assessment redesign is necessary. On dimensions where homogenization occurs (Cohesion Architecture, Structural Originality), assessment in AI-augmented environments may lose reliability, because inter-student differences are compressed by AI and the assessment risks measuring AI's standardization pattern rather than student ability. In contrast, dimensions resistant to homogenization (Perspective Plurality, Abstract--Concrete Oscillation) retain inter-student differences even in AI-augmented environments and therefore remain viable assessment targets. The homogenization profiles reported in this study provide an evidence-based guide for reselecting assessment criteria in the age of AI. In practice, large-scale AI-assisted assessment systems are already being deployed at major institutions; for example, Ritsumeikan AI-powered Assessment Solution (RaaS) at Ritsumeikan University processes thousands of student essays per semester. The integration of dimension-specific homogenization profiles into such systems could enable assessors to weight homogenization-resistant dimensions more heavily, thereby preserving assessment validity in AI-augmented environments.

\subsection{Limitations}
\label{subsec:limitations}

This study has several limitations. First, the H+AI conditions were generated computationally rather than through actual student--AI interaction. In this study, existing student essays were submitted to the AI via API, and augmented texts were obtained in a single turn. In practice, students improve essays through multi-turn dialogue with AI, selectively accepting or rejecting AI suggestions. This iterative process is not captured by the present design. In actual interaction, students' metacognitive judgment may serve a filtering function that mitigates homogenization. Conversely, multi-turn interaction could induce gradual convergence toward AI structural patterns. The directionality of this effect requires empirical investigation.

Second, the study is limited to English essays on two topics, constraining generalizability. Topic-level analysis confirmed genre-dependent effects on Abstract--Concrete Oscillation and Perspective Plurality. The finding that homogenization patterns differ between explanatory and argumentative essays warrants caution in generalizing to other languages, genres, and educational levels. In particular, AI structural patterns and student writing practices may differ in non-English languages such as Japanese, necessitating cross-linguistic replication studies.

Third, the results are dependent on the specific AI model used. Whether models with different architectures and training data produce the same homogenization patterns has not been verified. Determining whether homogenization is attributable to the characteristics of a specific model or constitutes a structural tendency common to large language models requires cross-model comparison studies.

Fourth, only three prompt conditions were tested, and they do not reflect the full range of student AI use patterns in practice. Students employ far more diverse prompts than the three conditions defined in this study, including multi-turn dialogue, paragraph-level revision requests, and information augmentation requests. Whether the finding that ``prompt specificity controls homogenization'' generalizes to this diversity of interaction modes remains an open question for future research.

%% file: Authors/documents/6_conclusion.tex

\section{Conclusion}
\label{sec:conclusion}

This study examined the impact of AI augmentation on both the quality and structural diversity of student writing by comparing the means and variances of six structural dimensions across 6,875 essays in five conditions.

Three main findings emerged. First, a Quality-Homogenization Tradeoff was confirmed across all conditions: AI augmentation substantially improved quality while simultaneously homogenizing specific dimensions. Second, homogenization was dimension-specific; Perspective Plurality and Abstract--Concrete Oscillation were diversified rather than homogenized. Third, even with the same AI model, prompt specificity reversed homogenization into diversification on Argument Depth, demonstrating that homogenization is a function of interaction design.

Future work should examine homogenization patterns in actual multi-turn student--AI interaction, validate the agreement between human expert evaluation and LLM-based evaluation, and investigate generalizability across multiple AI models, languages, and genres.

%% file: refs.bib
@article{1,
  author    = {Kofinas, A. K. and Tsay, C. H.-H. and Pike, D.},
  title     = {The impact of generative {AI} on academic integrity of authentic assessments within a higher education context},
  journal   = {British Journal of Educational Technology: Journal of the Council for Educational Technology},
  year      = {2025},
  volume    = {56},
  number    = {6},
  pages     = {2522--2549},
  doi       = {10.1111/bjet.13585}
}

@article{2,
  author    = {Inoshita, K. and Omura, M. and Yamanaka, T. and Maeda, G. and Tsuji, K.},
  title     = {Argument Rarity-based Originality Assessment for {AI}-assisted writing},
  journal   = {arXiv [cs.CL]},
  year      = {2026},
  doi       = {10.48550/arXiv.2602.01560}
}

@article{3,
  author    = {Hollan, J. and Hutchins, E. and Kirsh, D.},
  title     = {Distributed cognition: toward a new foundation for human-computer interaction research},
  journal   = {ACM Transactions on Computer-Human Interaction},
  year      = {2000},
  volume    = {7},
  number    = {2},
  pages     = {174--196},
  doi       = {10.1145/353485.353487}
}

@book{4,
  author    = {Maturana Rumesin, H. and Varela, F. J.},
  title     = {Autopoiesis and Cognition: The Realization of the Living},
  series    = {Boston Studies in the Philosophy and History of Science},
  year      = {1991},
  doi       = {10.1007/978-94-009-8947-4}
}

@article{5,
  author    = {McGuire, J. and De Cremer, D. and Van de Cruys, T.},
  title     = {Establishing the importance of co-creation and self-efficacy in creative collaboration with artificial intelligence},
  journal   = {Scientific Reports},
  year      = {2024},
  volume    = {14},
  number    = {1},
  pages     = {18525},
  doi       = {10.1038/s41598-024-69423-2}
}

@article{6,
  author    = {Steiss, J. and Tate, T. and Graham, S. and Cruz, J. and Hebert, M. and Wang, J. and others},
  title     = {Comparing the quality of human and {ChatGPT} feedback of students' writing},
  journal   = {Learning and Instruction},
  year      = {2024},
  volume    = {91},
  number    = {101894},
  doi       = {10.1016/j.learninstruc.2024.101894}
}

@article{7,
  author    = {Shi, H. and Chai, C. S. and Zhou, S. and Aubrey, S.},
  title     = {Comparing the effects of {ChatGPT} and automated writing evaluation on students' writing and ideal {L2} writing self},
  journal   = {Computer Assisted Language Learning},
  year      = {2025},
  pages     = {1--28},
  doi       = {10.1080/09588221.2025.2454541}
}

@article{8,
  author    = {Zhang, K.},
  title     = {Enhancing critical writing through {AI} feedback: A randomized control study},
  journal   = {Behavioral Sciences},
  year      = {2025},
  volume    = {15},
  number    = {5},
  pages     = {600},
  doi       = {10.3390/bs15050600}
}

@article{9,
  author    = {Meyer, J. and Jansen, T. and Schiller, R. and Liebenow, L. W. and Steinbach, M. and Horbach, A. and others},
  title     = {Using {LLMs} to bring evidence-based feedback into the classroom: {AI}-generated feedback increases secondary students' text revision, motivation, and positive emotions},
  journal   = {Computers and Education: Artificial Intelligence},
  year      = {2024},
  volume    = {6},
  number    = {100199},
  doi       = {10.1016/j.caeai.2023.100199}
}

@article{10,
  author    = {Connell Pensky, A. E. and Usdan, J. H. and Chang, H.},
  title     = {Generative {AI}'s impact on graduate student professional writing productivity and quality},
  journal   = {International Journal of Artificial Intelligence in Education},
  year      = {2025},
  volume    = {35},
  number    = {6},
  pages     = {4057--4082},
  doi       = {10.1007/s40593-025-00528-z}
}

@article{11,
  author    = {Burstein, J. and Braden-Harder, L. and Chodorow, M. and Hua, S. and Kaplan, B. and Kukich, K. and others},
  title     = {Computer analysis of essay content for automated score prediction: A prototype automated scoring system for {GMAT} analytical writing assessment essays},
  journal   = {ETS Research Report Series},
  year      = {1998},
  volume    = {1998},
  number    = {1},
  pages     = {i--67},
  doi       = {10.1002/j.2333-8504.1998.tb01764.x}
}

@article{12,
  author    = {Hussein, M. A. and Hassan, H. and Nassef, M.},
  title     = {Automated language essay scoring systems: a literature review},
  journal   = {PeerJ. Computer Science},
  year      = {2019},
  volume    = {5},
  number    = {e208},
  doi       = {10.7717/peerj-cs.208}
}

@inproceedings{13,
  author    = {Taghipour, K. and Ng, H. T.},
  title     = {A Neural Approach to Automated Essay Scoring},
  booktitle = {Proceedings of the 2016 Conference on Empirical Methods in Natural Language Processing},
  year      = {2016},
  pages     = {1882--1891},
  doi       = {10.18653/v1/D16-1193}
}

@inproceedings{14,
  author    = {Naismith, B. and Mulcaire, P. and Burstein, J.},
  title     = {Automated evaluation of written discourse coherence using {GPT-4}},
  booktitle = {Proceedings of the 18th Workshop on Innovative Use of NLP for Building Educational Applications (BEA 2023)},
  year      = {2023},
  pages     = {394--403},
  doi       = {10.18653/v1/2023.bea-1.32}
}

@inproceedings{15,
  author    = {Padmakumar, V. and He, H.},
  title     = {Does writing with language models reduce content diversity?},
  booktitle = {Proceedings of the 41st International Conference on Machine Learning},
  year      = {2023},
  doi       = {10.48550/arXiv.2309.05196}
}

@article{16,
  author    = {Moon, K. and Green, A. E. and Kushlev, K.},
  title     = {Homogenizing effect of large language models ({LLMs}) on creative diversity: An empirical comparison of human and {ChatGPT} writing},
  journal   = {Computers in Human Behavior: Artificial Humans},
  year      = {2025},
  volume    = {6},
  number    = {100207},
  doi       = {10.1016/j.chbah.2025.100207}
}

@inproceedings{17,
  author    = {Anderson, B. R. and Shah, J. H. and Kreminski, M.},
  title     = {Homogenization effects of large language models on human creative ideation},
  booktitle = {Creativity and Cognition},
  year      = {2024},
  pages     = {413--425},
  doi       = {10.1145/3635636.3656204}
}

@article{18,
  author    = {Reviriego, P. and Conde, J. and Merino-G\'{o}mez, E. and Mart\'{\i}nez, G. and Hern\'{a}ndez, J. A.},
  title     = {Playing with words: Comparing the vocabulary and lexical diversity of {ChatGPT} and humans},
  journal   = {Machine Learning With Applications},
  year      = {2024},
  volume    = {18},
  number    = {100602},
  doi       = {10.1016/j.mlwa.2024.100602}
}

@article{19,
  author    = {Guo, Y. and Shang, G. and Clavel, C.},
  title     = {Benchmarking linguistic diversity of Large Language Models},
  journal   = {Transactions of the Association for Computational Linguistics},
  year      = {2025},
  volume    = {13},
  pages     = {1507--1526},
  doi       = {10.1162/tacl.a.47}
}

@article{20,
  author    = {Xu, W. and Jojic, N. and Rao, S. and Brockett, C. and Dolan, B.},
  title     = {Echoes in {AI}: Quantifying lack of plot diversity in {LLM} outputs},
  journal   = {Proceedings of the National Academy of Sciences of the United States of America},
  year      = {2025},
  volume    = {122},
  number    = {35},
  pages     = {e2504966122},
  doi       = {10.1073/pnas.2504966122}
}

@inproceedings{21,
  author    = {Guo, Y. and Shang, G. and Vazirgiannis, M. and Clavel, C.},
  title     = {The curious decline of linguistic diversity: Training language models on synthetic text},
  booktitle = {Findings of the Association for Computational Linguistics: NAACL 2024},
  publisher = {Association for Computational Linguistics},
  year      = {2024},
  pages     = {3589--3604},
  doi       = {10.18653/v1/2024.findings-naacl.228}
}

@misc{22,
  author    = {OpenAI},
  title     = {{GPT-5} Model},
  note      = {\url{https://developers.openai.com/api/docs/\ models/gpt-5} Accessed 26.03.20}
}

@article{23,
  author    = {Zheng, L. and Chiang, W.-L. and Sheng, Y. and Zhuang, S. and Wu, Z. and Zhuang, Y. and others},
  title     = {Judging {LLM}-as-a-judge with {MT-bench} and {Chatbot Arena}},
  journal   = {arXiv [cs.CL]},
  year      = {2023},
  doi       = {10.48550/arXiv.2306.05685}
}

@article{24,
  author    = {Tang, X. and Chen, H. and Lin, D. and Li, K.},
  title     = {Harnessing {LLMs} for multi-dimensional writing assessment: Reliability and alignment with human judgments},
  journal   = {Heliyon},
  year      = {2024},
  volume    = {10},
  number    = {14},
  pages     = {e34262},
  doi       = {10.1016/j.heliyon.2024.e34262}
}

@article{25,
  author    = {Kim, Y.},
  title     = {Automated essay scoring with {GPT-4} for a local placement test: Investigating prompting strategies, intra-rater reliability, and alignment with human scores},
  journal   = {TESOL Quarterly},
  year      = {2025},
  volume    = {59},
  number    = {S1},
  pages     = {S318--S329},
  doi       = {10.1002/tesq.3405}
}

@article{26,
  author    = {Quah, B. and Zheng, L. and Sng, T. J. H. and Yong, C. W. and Islam, I.},
  title     = {Reliability of {ChatGPT} in automated essay scoring for dental undergraduate examinations},
  journal   = {BMC Medical Education},
  year      = {2024},
  volume    = {24},
  number    = {1},
  pages     = {962},
  doi       = {10.1186/s12909-024-05881-6}
}

@article{27,
  author    = {Liu, Y. and Qi, H. and Lu, X.},
  title     = {Enhancing {GPT}-based automated essay scoring: the impact of fine-tuning and linguistic complexity measures},
  journal   = {Computer Assisted Language Learning},
  year      = {2025},
  pages     = {1--20},
  doi       = {10.1080/09588221.2025.2518430}
}

@article{28,
  author    = {Wataoka, K. and Takahashi, T. and Ri, R.},
  title     = {Self-preference bias in {LLM}-as-a-judge},
  journal   = {arXiv [cs.CL]},
  year      = {2024},
  doi       = {10.48550/arXiv.2410.21819}
}

@article{29,
  author    = {Ye, J. and Wang, Y. and Huang, Y. and Chen, D. and Zhang, Q. and Moniz, N. and others},
  title     = {Justice or prejudice? Quantifying biases in {LLM}-as-a-Judge},
  journal   = {arXiv [cs.CL]},
  year      = {2024},
  doi       = {10.48550/arXiv.2410.02736}
}

@article{30,
  author    = {Bevilacqua, M. and Oketch, K. and Qin, R. and Stamey, W. and Zhang, X. and Gan, Y. and others},
  title     = {When automated assessment meets automated content generation: Examining text quality in the era of {GPTs}},
  journal   = {ACM Transactions on Information Systems},
  year      = {2025},
  volume    = {43},
  number    = {2},
  pages     = {1--36},
  doi       = {10.1145/3702639}
}

@misc{31,
  author    = {Burleigh, L.},
  title     = {{AIDE}: {AI} Detection for Essays Dataset},
  year      = {2024},
  note      = { \url{https://www.kaggle.com/lburleigh/tla-lab-ai-detection-for-\ essays-aide-dataset Accessed 26.01.31}}
}

@misc{32,
  author    = {OpenAI},
  title     = {{GPT-5} mini Model},
  note      = {\url{https://developers.openai.com/api/docs\ /models/gpt-5-mini} Accessed 26.03.20}
}
